\documentclass[a4paper,conference]{IEEEtran}

\ifCLASSINFOpdf
\else
\fi

\usepackage{times}
\usepackage{epsfig}
\usepackage{graphicx}
\usepackage{amsmath}
\usepackage{amssymb}
\usepackage{enumitem}
\usepackage{multirow}
\usepackage[dvipsnames]{xcolor}
\usepackage{comment}
\usepackage[normalem]{ulem}
\usepackage{xspace}
\usepackage{booktabs,makecell}
\usepackage{hyperref}
\makeatletter

\usepackage{tikz}
\usetikzlibrary{arrows,shapes,calc,matrix,fit,backgrounds}
\usepackage{pgfplots}
\usepackage{pgfplotstable}
\pgfplotsset{compat=1.9}
\usepgfplotslibrary{fillbetween}

\usepackage{xstring}

\usepgfplotslibrary{external}

\IfBeginWith*{\jobname}{fig/extern/}{\finalcopy}{}

\tikzset{every mark/.append style={solid}}
\pgfplotsset{
	grid=both, width=\columnwidth, try min ticks=5,
	every axis/.append style={font=\scriptsize},
	every axis plot/.append style={thick,mark=none,mark size=1.2,tension=0.18},
	legend cell align=left, legend style={fill opacity=0.8},
}

\pgfplotsset{
	dash/.style={mark=o,dashed,opacity=0.7},
	dott/.style={mark=o,dotted,opacity=0.7},
}

\def\l1{\ensuremath{\ell_1}\xspace}
\def\l2{\ensuremath{\ell_2}\xspace}

\def\xssp{\hspace{2pt}}

\newcommand{\cD}{\mathcal{D}}

\newcommand{\cT}{\mathcal{T}}

\newcommand{\cX}{\mathcal{X}}
\newcommand{\cY}{\mathcal{Y}}

\makeatletter
\DeclareRobustCommand\onedot{\futurelet\@let@token\@onedot}
\def\@onedot{\ifx\@let@token.\else.\null\fi\xspace}
\def\eg{\emph{e.g}\onedot} 
\def\ie{\emph{i.e}\onedot}

\def\etal{\emph{et al}\onedot}
\makeatother

\begin{document}
%
\title{Edge Augmentation for Large-Scale Sketch Recognition without Sketches}

\author{\IEEEauthorblockN{Nikos Efthymiadis}
\IEEEauthorblockA{VRG, Faculty of Electrical Engineering \\
Czech Technical University in Prague\\
\small Email: efthynik@fel.cvut.cz}
\and
\IEEEauthorblockN{Giorgos Tolias}
\IEEEauthorblockA{VRG, Faculty of Electrical Engineering \\
Czech Technical University in Prague\\
\small Email: giorgos.tolias@cmp.felk.cvut.cz}
\and
\IEEEauthorblockN{Ondřej Chum}
\IEEEauthorblockA{VRG, Faculty of Electrical Engineering \\
Czech Technical University in Prague\\
\small Email: chum@cmp.felk.cvut.cz}}


%


\maketitle

\begin{abstract}

This work addresses scaling up the sketch classification task into a large number of categories. 
Collecting sketches for training is a slow and tedious process that has so far precluded any attempts to large-scale sketch recognition. We overcome the lack of training sketch data by exploiting labeled collections of natural images that are easier to obtain.
To bridge the domain gap we present a novel augmentation technique that is tailored to the task of learning sketch recognition from a training set of natural images. Randomization is introduced in the  parameters of edge detection and edge selection. Natural images are translated to a pseudo-novel domain called ``randomized Binary Thin Edges'' (rBTE), which is used as a training domain instead of  natural images. 
The ability to scale up is demonstrated by training CNN-based sketch recognition of more than 2.5 times larger number of categories than used previously.
For this purpose, a dataset of natural images from 874 categories is constructed by combining a number of popular computer vision datasets. The categories are selected to be suitable for sketch recognition. To estimate the performance, a subset of 393 categories with sketches is also collected.
\end{abstract}


%
\IEEEpeerreviewmaketitle

\section{Introduction}
\label{sec:intro}
Free-hand drawings, or \emph{sketches}, have been a long-lasting means of human communication and expression. 
Nowadays, the prevalence of digital devices equipped with touch-screens has given free-hand sketches additional
roles in a number of educational, business or leisure activities. 
As a result, computer vision research related to sketches has flourished in a variety of tasks including synthesis~\cite{he18}, perceptual grouping~\cite{yzf+20}, sketch-based image retrieval~\cite{bc15,rtc18,wkl15}, and sketch recognition~\cite{zlz+16,jfy+20,ldl+19}. 
In this paper, we focus on the task of sketch recognition, \ie how to classify sketches into specific categories.
In particular, we target a realistic application scenario, where the number of classes is as large as possible.
Prior work commonly keeps the number of classes relatively low. The reason for that is simple - lack of training data. Since annotation effort for sketch recognition includes sketch drawing, this activity becomes prohibitively expensive. For example, the \emph{Sketchy} dataset~\cite{sbh+16} required 3,921 hours of sketching for 125 categories. In order to obtain a seven times larger dataset, one would need {\bf over 13 human-years} of sketch drawing (40 hours a week, 52 weeks a year). Therefore, the task of large-scale sketch recognition requires methods much less demanding on the training data.

To allow scalability in the number of classes, we propose a method that trains a deep network classifier without requiring a single sketch during the training. Instead, only natural images with their labels are used to train the classifier. The method exploits the fact that human-drawn sketches often represent the 2D shape of depicted objects or of their parts. The sketch domain is approximated by detected 2D shapes of objects in natural images. In the following, terms \emph{natural images}, \emph{RGB images}, or simply \emph{images} are used interchangeably.

\begin{figure}[t]
\vspace{-5pt}
\begin{center}
\input{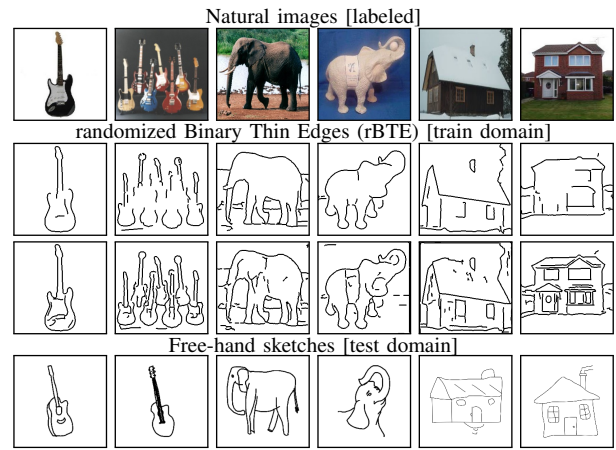}
\vspace{-5pt}
\caption{The goal of this work is to recognize sketches at test time (bottom row) without using any sketches at training time. Labeled natural images (top row) are transformed to rBTEs with different level of details (the two middle rows are two instances of rBTE per natural image, thickened for visualization) to bridge the domain gap. Combined with geometric augmentations, the transformed dataset is used to train a deep network for sketch recognition.
\label{fig:teaser}
\vspace{-15pt}}
\end{center}
\vspace{-9pt}
\end{figure}

A novel edge augmentation technique is used to map natural images to a  pseudo-novel domain called randomized Binary Thin Edges (rBTE). This augmentation procedure randomly selects an edge detector and an edge selection strategy in order to generate a sketch-like output with different level of details (see Figure~\ref{fig:teaser}) and is combined with random geometric augmentations. 

Sketch recognition is a standard and well defined task and, at the same time, collections of annotated images are available. It is possible to cast sketch recognition as a single-source domain generalization with natural images as the source domain. However, we show that a specific approach exploiting the specifics of natural images and sketches brings a relative recognition-rate improvement of more than 20\% over unnecessarily generic single-source domain generalization approaches.

The proposed approach is a general augmentation scheme that can include any modern image-to-edge or image-to-sketch method. In this work we demonstrate its potential by using edge detectors~\cite{dz13,sz17,jsm+19} trained on an extremely limited amount of non-sketch data, \ie 200 natural images. Methods such as~\cite{bcs+21, sps+18} are trained on sketches, and methods such as~\cite{lzd13} require more data. rBTEs form a rich training set allowing to train, without a single sketch, a CNN-based sketch classifier, which is the main contribution of this work.

Sketch synthesis is a popular task~\cite{izz+17, jaf16,gpj+14,zpt+17, zsq+19} in which sketches are generated from images. However, it is not applicable to the setup that this work explores since these approaches cannot work without training sketches.

To evaluate the proposed approach we introduce Im4Sketch, a dataset for large-scale sketch recognition without sketches for training. It consists of 1,007,878 natural images labeled into 874 classes used to train the sketch classifier. Testing is performed on 80,582 sketches coming from 393 classes that are a subset of the training classes. The dataset is a composition of existing popular image and sketch datasets, namely ImageNet~\cite{rds+15}, DomainNet (DN)~\cite{pbx19}, Sketchy~\cite{sbh+16}, PACS~\cite{lys+17}, and TU-Berlin~\cite{eha12}. The classes are selected, so that classification by shape is meaningful. For example, ImageNet categories ``Indian elephant'' and ``African elephant'' are merged into category ``Elephant''. The dataset is described in detail in Section~\ref{sec:dataset}.

To the best of our knowledge, this is the first work that delivers sketch recognition of the order of over 800 categories.
The dataset with the second largest number of classes for sketch recognition is DomainNet \cite{pbx19} with 345 classes, \ie more than 2.5 times smaller.

\section{Related work}
\label{sec:related}
\begin{figure*}[t]
\begin{center}
\includegraphics[width=0.90\linewidth]{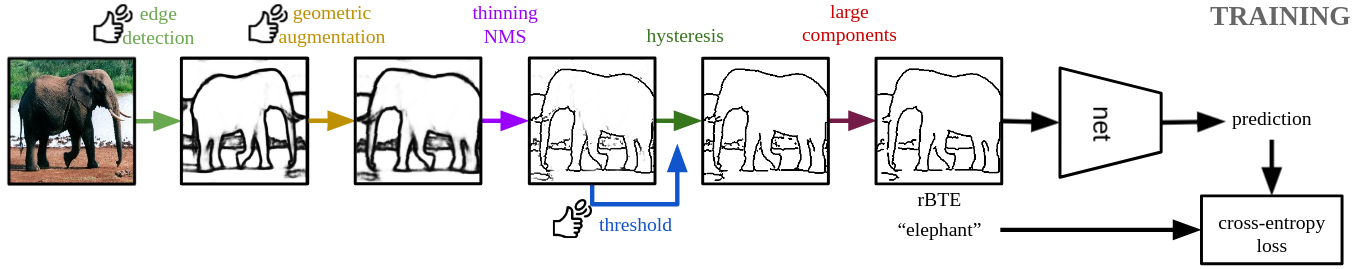}
\vspace{-10pt}
\caption{Overview of the training pipeline. 
Natural images are transformed into rBTEs, which are used with class labels to train a network classifier with cross-entropy loss. The obtained network is used to classify free-hand sketches into the object categories. 
\label{fig:overview}
}
\vspace{-15pt}
\end{center}
\end{figure*}

In this section, we review the prior work on three tasks that are relevant to the focus of this work, namely sketch recognition, sketch-based image retrieval, and domain generalization.

\paragraph{Sketch recognition}
The crowd-sourced free-hand sketch dataset by Eitz~\etal~\cite{eha12} is the first large-scale dataset on the domain of sketches. 
Early approaches~\cite{st14,eha12} focus on adapting hand-crafted features and encodings, such as SIFT~\cite{l04}, its variants, Fisher vectors~\cite{pd07}, and SVM classifiers to solve the task. 
The application of deep network classifiers was stimulated by the enormous effort invested in annotating sketches.
The \emph{Sketch-a-Net}~\cite{yys+15} approach demonstrates recognition performance surpassing human recognition ability. This is achieved with a tailored architecture and a labeled training set of sketches whose size is in the order of $10^6$.
In the recent work of Qi~\etal~\cite{jfy20}, the combination of deep and hand-crafted features exhibits very good results.
Some approaches~\cite{xjb19,lzs18} exploit the additional information of stroke order, when available in the input, to further improve the recognition accuracy.
We assume that this information is not available and deal with the more general problem.

The scarcity of training data in the domain of sketches is handled by some approaches by combining sketches and natural images during the training.
Hua \etal~\cite{zlz+16} attempt to automatically learn the shared latent structures that exist between sketch images and natural images.
Zhang~\etal~\cite{zlm+19} transfer the knowledge of a network learned on natural images to a sketch network. 
In both these methods the training set consists of both natural images and sketches. 
Even though these approaches are valuable in a few-shot setup, where only a few labeled sketches per category are available, the setup with no sketches has not been well studied before with a focus on the sketch domain.
An exception is the recent work of Lamb~\etal~\cite{lov+20} where the \emph{SketchTransfer} task is presented. Even though their work  explores the setup of no available sketches too, promising results are achieved only when unlabeled sketches are available during the training.
The authors conclude that the low resolution images of the benchmark is a limitation. Therefore, in our work we use benchmarks with higher resolution images that are more realistic.

\paragraph{Sketch-based image retrieval}
Classical approaches use edge detection on natural images to bridge the domain gap and then handle both domains with hand-crafted descriptors or matching~\cite{rc13,bc15,sbo15,pm14,tc17}.
Deep learning methods mainly follow a different path. 
A two branch architecture is used~\cite{brp+16b,sdm17,yls+16}, with a different branch per domain, where the expectation is to bridge the domain gap based on large amounts of training data with cross-domain labeling~\cite{sbh+16}.
If learning is involved, the most realistic setup is the zero-shot sketch-based image retrieval~\cite{drd+19}; 
which is a challenging task, that is related to, but different from, ours. 
Radenovic \etal~\cite{rtc18} avoid cross-modal annotation by relying on training labels of natual images and using edge detection to bridge the domain gap. Their work focuses on learning shape similarity and does not attempt to directly generalize to category level recognition.

\paragraph{Domain generalization}
The most common approach for domain generalization is invariant feature learning, based on the theoretical results of Ben-David \etal ~\cite{bbc+07}. 
Representative approaches include kernel-based invariant feature learning by minimizing domain dissimilarity\cite{mbs13}, multi-task autoencoders that transform the original image to other related domains, domain classifiers as adversaries to match the source domain distributions in the feature space~\cite{gua+16,ltg+18}, and cross-domain non-contrastive learning as regularization~\cite{SelfReg}.

Some methods specialize for single-source domain generalization. Examples include hard example generation in virtual target domains ~\cite{ADA}, style transfer using auxiliary datasets~\cite{yzz+19}, and adversarial domain augmentation ~\cite{qzp20}. Narayanan \etal ~\cite{nrk21} argue that the shock graph of the contour map of an image is a complete representation of its shape content and use a Graph Neural Network as their model. Wang \etal ~\cite{Wlq+21} propose a style-complement module to create synthetic images from distributions that are complementary to the source domain.

Data augmentation techniques are commonly used for domain generalization. Zhou \etal ~\cite{zyh+20} synthesize data from pseudo-novel domains under semantic consistency by using a data generator. Mancini \etal ~\cite{mar+20} use mixup~\cite{zcd+18} to combine different source domains. Carlucci \etal ~\cite{JiGen} train a model to solve jigsaw puzzles in a self-supervised manner in addition to the standard classification loss to improve the generalization ability of the model. 

\section{Task formulation}
\label{sec:task}
In this section, we define the task and relate it to existing computer vision tasks. We follow the notation of transfer learning~\cite{wkt+16} and domain adaptation~\cite{c17} literature.

A domain $\cD$ is an ordered pair $\cD=(\cX, P(X))$ composed of a space of input examples $\cX$ and a marginal probability distribution $P(X)$, where $X$ is a random variable valued in $\cX$. 
A task $\cT=(\cY, P(Y|X))$ is defined by a label space $\cY$ and the conditional probability distribution $P(Y|X)$, where $Y$ is a random variable with values in $\cY$. 

In the problem, two domains are considered: the \emph{target} domain $\cD^t=(\cX^t, P(X^t))$ of sketches  and the \emph{source} domain $\cD^s=(\cX^s, P(X^s))$ of natural images, with tasks $\cT^t=(\cY^t, P(Y^t|X^t))$ and $\cT^s=(\cY^s, P(Y^s|X^s))$ respectively. The goal is to learn a predictor $f: \cX^t \rightarrow \cY^t$ for the target domain without having access to any examples from that domain.

The input spaces of both the domains, target and source respectively, are images (RGB, fixed size), thus $\cX^s=\cX^t$. The same categories are to be recognized in the two domains, \ie the label spaces are also identical $\cY^s=\cY^t$. However, the marginal distributions are significantly different, \ie $P(X^s)\neq P(X^t)$.
In this work, we advocate for bridging the domain gap by constructing a transformation $T: \cX \rightarrow \cX$ so that $P(T(X^s))\approx P(X^t)$.
With such a transformation, an approximation of $P(Y^t|X^t)$ in the form of $P(Y^s|T(X^s))$ can be learned with labeled examples from the source domain of natural images. 
In this work, we focus on designing the transformation based on prior knowledge about the two domains, see Section~\ref{sec:method}.

\paragraph{Relation to domain adaptation}
In the domain adaptation task, similarly to our problem, $\cY^s=\cY^t$ and $P(X^s)\neq P(X^t)$. The main difference is that in domain adaptation, some examples from the target domain are available, either labeled or unlabeled for supervised or unsupervised domain adaptation respectively.
For example, in unsupervised domain adaption from natural images to sketches, a labeled dataset of natural images is available, together with unlabeled sketch examples. The goal is to obtain a predictor for sketches.

\paragraph{Relation to domain generalization}
The task of domain generalization is the closest one to our task. 
The essential difference is that most domain generalization methods are either unusable or they under-perform in the single-source task. Also domain generalization targets to perform well in every possible target domain. 
The domain label of each example is used by most approaches as additional supervision. It holds that $P(X^s_j)\neq P(X^t)$ and $\cY^s_j=\cY^t$ for $j=1\ldots d$, where $d$ is the number of source domains.
A sketch recognition example is the case where labeled datasets exist for the domain of natural images, artworks, and cartoons. The goal is to obtain a predictor for free-hand sketches. In contrast to our setup, the multiple domains allow for learning a domain invariant predictor, while in our task, exploiting prior knowledge is the only way to proceed.

\paragraph{Relation to attributed-based zero-shot learning}
Zero-shot learning in visual applications is the following. The input spaces are the same $\cX^s=\cX^t$, but the marginal distributions are different $P(X^s)\neq P(X^t)$. The label spaces are disjoint $\cY^s \cap \cY^t = \emptyset$, and, therefore, the tasks are different too, \ie $\cT^s \neq \cT^t$. 
The similarity to our task is that there are no input examples of the target domain during the learning. 
There is additional information, though, in the form of attributes. Each class, from both domains, is described by an attribute vector, whose dimensions correspond to high-level semantically meaningful properties~\cite{lnh13}. 
This information is used to transfer knowledge from one domain to the other.

\section{Method}
\label{sec:method}

\begin{figure}[t]
\vspace{-5pt}
\begin{center}
\input{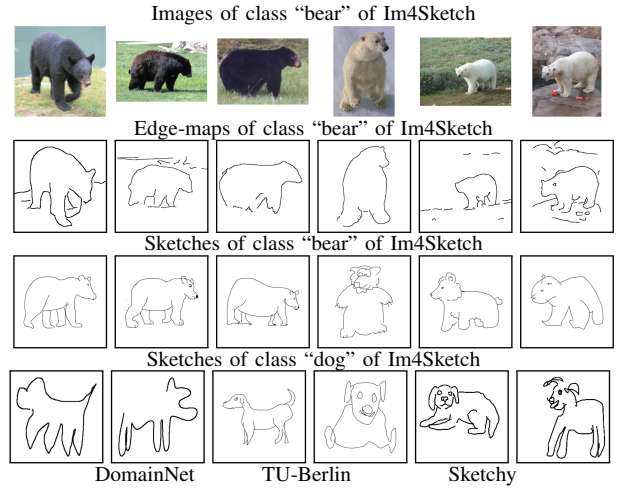}
\vspace{-5pt}
\caption{Image and sketch examples from the Im4Sketch dataset. Class bear contains the original ImageNet classes ``American Black Bear" and ``Ice Bear" whose shape is indistinguishable. Sketches are collected from original datasets with different level of detail as in the case of class ``dog''; original datasets are shown at the bottom.
\label{fig:dataset}
\vspace{-20pt}
}
\end{center}
\end{figure}

In this section, we describe the proposed method, the details of the construction of the rBTE domain, as well as the implementation details of the training.

\subsection{Deep network classifier}
Predictor $f: \cX^t \rightarrow \cY^t$ is a deep convolutional neural network that takes as input a sketch and predicts the class. 
It is possible to obtain the predictor by minimizing empirical risk $\frac{1}{n} \sum_{i=1}^{n} \ell(f(x_i), y_i)$, with $(x_i,y_i) \in \cX^t \times \cY^t$, when sketches are available during training. 
Instead, in this work, the predictor is obtained by minimizing $\frac{1}{n} \sum_{i=1}^{n} \ell(f(T(x_i), y_i)$, with $(x_i,y_i) \in \cX^s \times \cY^s$.

\subsection{Edge Augmentation}
A natural image $x$ is transformed to $T(x)$, called rBTE, through a sequence of randomized transformations that are described in the following.

\paragraph{Edge detection} is performed to map the input image to an edge-map with values in $[0,1]$. 
Various edge detectors are used to extract the edges: The \emph{Structures Edges} (SE)~\cite{dz13}, the \emph{Holistically-Nested Edge Detection}\footnote{A reimplementation is used \url{https://github.com/sniklaus/pytorch-hed}}~\cite{sz17}, and the \emph{Bi-Directional Cascade Network}~\cite{jsm+19}.
While SE uses random-forest classifier, the other two are DNN based.
Using multiple detectors, the size of available training examples is expanded, in this case by a factor of 3. 

All three edge detectors are originally trained with clean edge-maps obtained from ground truth segmentation masks, in particular the BSDS500~\cite{amf+11} segmentation dataset containing only 500 images, 200 of which are for training. The dataset consists of natural images, and the ground truth masks are obtained by manual labeling. 
This is an additional labeled dataset, not including free-hand sketches of objects, that is indirectly used in our work to improve sketch recognition. 
Even though not considered in this work, one could possible make better use of this dataset even during learning predictor $f$ to obtain further improvements. 

\paragraph{Geometric augmentations}
Edge-maps are geometrically transformed with a set of commonly used geometric augmentations for CNNs: 
Zero-padding to obtain square edge-maps, re-sample to 256$\times$256, rotation by angle uniformly sampled in $[-5,5]$ degrees, crop with relative size uniformly sampled in $[0.8,1.0]$ and aspect ratio in $[\frac{3}{4},\frac{4}{3}]$ resized to 224$\times$224, and finally horizontal flip with a probability of 0.5.

\paragraph{Thinning - NMS}
Edge thinning is performed by non-maximum suppression that finds the locations with the sharpest change of the edge strength. 
All values that are not larger than the neighboring values along the positive and negative gradient directions are suppressed. 
This is a standard process which is also part of the well known Canny detector~\cite{c86} and commonly used as post-processing for the SE detector.

\paragraph{Hysteresis thresholding}
Hysteresis thresholding, which is another common step of the Canny~\cite{c86} detector, that transforms the input to a binary images using two thresholds (low and high).
Pixels whose value is above (below) the high (low) threshold are (are not) considered edges, while pixels with values between the two thresholds are considered as edges only if they belong to a connected-component containing an edge pixel.
Setting a fixed threshold that operates well for the large range of images and objects considered in this work is not possible.
Instead, a threshold $t$ is estimated by standard approaches that perform binarization of grayscale images, and set the low and high thresholds to $0.5t$ and $1.5t$, respectively. 
The thresholding approach is randomly chosen from a candidate pool comprising methods of Otsu~\cite{o79}, Yen~\cite{ycc95}, Li~\cite{ll93}, Isodata~\cite{rc78}, and the mean approach~\cite{g93}.

\paragraph{Large connected-components}
In the last stage, pixels belonging to small (less than 10 pixels) connected components, estimated with 8-neighbor connectivity, are discarded.

\subsection{Testing} The relative size of sketches with respect to the image dimensions varies from dataset to dataset. In order to tackle this we perform inference in the following two ways. single-scale testing, which is the testing of each sketch in its original relative size and multi-scale testing. In multi-scale testing the sketch is cropped to its bounding box, padded to have aspect ratio 1:1 and then resized to be 90\%, 65\% and 45\% of the 224x224 network input size. The prediction of the three resized sketches are then ensembled by averaging before the softmax function. 

\subsection{Overview}
A visual overview of the pipeline is shown in Figure~\ref{fig:overview}.
The proposed approach uses the sequence of transformations on natural images during the learning stage, and trains a deep convolutional neural network with rBTEs.
Randomness is involved in the selection of the edge-map, in the geometric augmentation, and in the selection of the threshold estimation method. Each time an image participates in a batch, only one of the edge-maps and one of the threshold estimation methods is used, chosen with equal probability. 
A set of thorough ablations is presented in the experimental section by discarding parts of the overall pipeline.
During inference, a sketch is fed to the exact same network, after simply performing thinning. 

\subsection{Implementation details}

\paragraph{Training details}
ResNet-101~\cite{hzr+16} is used as the backbone network for our ablation study on Sketchy and for the core Im4Sketch experiments. 
The parameters of the network are initialized with the standard weights from training on ImageNet, \ie training with rBTEs starts with the network trained on ImageNet with RGB images. Adam optimizer is used with batch size equal to 64 for training on rBTEs.
The \emph{range test}~\cite{l17} is used to identify the initial learning rate. 
This process starts with a high learning rate and keeps decreasing it every 50 updates. The  initial learning rate is picked in the range of the steepest ascending accuracy on the training set. 
This method indicates a learning rate of $10^{-4}$ for all of our experiments.
The learning rate is decayed by a factor of 10 every 10 epochs for a total of 30 epochs.

\paragraph{Comparison with domain generalization methods}
For a fair comparison with the existing domain generalization methods we train a ResNet-18~\cite{hzr+16} with SGD optimizer with momentum 0.9, batch size  64 and learning rate $0.004$. We train for 30 epochs without a scheduler.

\section{The Im4Sketch dataset}
\label{sec:dataset}

\begin{table}[t]
\vspace{8pt}
\def\arraystretch{0.95}
\begin{center}
\footnotesize
\setlength\extrarowheight{1pt}
\begin{tabular}{@{\xssp}l@{\xssp}c@{\xssp}|@{\xssp}c@{\xssp}c@{\xssp}c@{\xssp}c@{\xssp}c@{\xssp}c@{\xssp}}
\hline
Dataset &
  Im4Sketch &
  ImageNet &
  DomainNet &
  Sketchy &
  PACS &
  TU-Berlin \\ \hline \hline
ImageNet  & 690 & -    & 162    & 99     & 4     & 141    \\
DomainNet & 345 & 162    & -    & 98     & 7     & 197    \\
Sketchy   & 125 & 99     & 98     & -    & 5     & 99     \\
PACS      & 7 & 4      & 7      & 5      &  -    & 7     \\
TU-Berlin & 223 & 141    & 197    & 99     & 7     & -   \\ \hline 
\end{tabular}
\end{center}

\vspace{-4pt}
\caption{Each row represents the original dataset as a component of Im4Sketch. The Im4Sketch column shows the number of final classes that exist in each dataset. The rest of the columns show the number of common final classes, after our merging, between the different datasets. ImageNet is with our merged version. TU-Berlin is without the deleted classes. \label{tab:datasets1}
\vspace{-5pt}
}
\vspace{-20pt}
\end{table}

We present a large-scale dataset with shape-oriented set of classes for image-to-sketch generalization called ``Im4Sketch''\footnote{Im4Sketch is publicly available on \url{http://cmp.felk.cvut.cz/im4sketch/}}.
It consists of a collection of natural images from 874 categories for training and validation, and sketches from 393 categories (a subset of natural image categories) for testing.

The images and sketches are collected from existing popular computer vision datasets. The categories are selected having shape similarity in mind, so that object with same shape belong to the same category. 

The natural-image part of the dataset is based on the ILSVRC2012 version of ImageNet (IN)~\cite{rds+15}. The original ImageNet categories are first merged according to the shape criteria. 
Object categories for objects whose shape, \eg how a human would draw the object, is the same are merged. For this step, semantic similarity of categories, obtained through WordNet~\cite{f98b} and category names, is used to obtain candidate categories for merging. Based on visual inspection of these candidates, the decision to merge the original ImageNet classes is made by a human. For instance, "Indian Elephant" and "African Elephant", or "Laptop" and "Notebook" are merged. An extreme case of merging is the new class “dog” that is a union of 121 original ImageNet classes of dog breeds.

In the second step, classes from datasets containing sketches are used. In particular,
DomainNet (DN)~\cite{pbx19}, Sketchy~\cite{sbh+16}, PACS~\cite{lys+17}, and TU-Berlin~\cite{eha12}.
Note that merging is not necessary for classes in these datasets, because the shape criteria are guaranteed since they are designed for sketches.
In this step, a correspondence between the merged ImageNet categories and categories of the other datasets is found. As in the merging step, semantic similarity is used to guide the correspondence search. A summary of the common classes per dataset pairs is shown in Table~\ref{tab:datasets1}.
Sketch categories that are not present in the merged ImageNet are added to the overall category set, while training natural images of those categories are collected from either DomainNet or Sketchy. In the end,
ImageNet is used for 690 classes, DomainNet for 183 classes, and Sketchy for 1 class, respectively.
An example of merging and mapping is shown in Figure~\ref{fig:dataset}.

Training images of Im4Sketch come from training sets of ImageNet and DomainNet(rel), and from 80\% of training set of Sketchy. The validation set is obtained from the validation set of ImageNet, the test set of DomainNet(rel) and 20\% of the training set of Sketchy.
To avoid large imbalance when collecting images from ImageNet we keep at most 1350 images per class, chosen uniformly from all corresponding original ImageNet classes. 

Almost all sketch categories from the four datasets are covered in Im4Sketch. 
We exclude 27 classes of the TU-Berlin dataset in order to either avoid class conflicts, \eg ''flying bird'' and ''standing bird'', or because we are unable to map them to any existing category with natural images in another dataset, \eg ''sponge bob''.
All sketches assigned to any of the final set of categories are used to form the Im4Sketch test set, with an exception for Sketchy and DomainNet where we keep only the sketches from the test set; see more details in Table~\ref{tab:datasets2}.

\begin{table}[t]
\vspace{8pt}
\def\arraystretch{0.85}
\begin{center}
\footnotesize
\setlength\extrarowheight{2pt}
\begin{tabular}{@{\xssp}l@{\xssp}c@{\xssp}c@{\xssp}c@{\xssp}c|@{\xssp}c@{\xssp}c@{\xssp}c}
\hline
 & \multicolumn{4}{c}{Original Datasets}  & \multicolumn{3}{c}{Im4Sketch}\\ \hline
Dataset    & Train    & Val   & Test   & Classes & Train    & Val   & Test\\ \hline \hline
IN         & 1,281,167  & 50,000 & 100,000 & 1,000  & 885,946   & 34,500 & 0 \\
DN Real    & 120,906   & 0     & 52,041  & 345  & 61,039    & 26,303 & 0 \\
DN Qdr     & 120,750   & 0     & 51,750  & 345 & 0   & 0    & 51,750 \\
Sketchy Im & 11,250    & 0     & 1,250   & 125 & 72       & 18    & 0  \\
Sketchy Sk & 68,418    & 0     & 7,063   & 125 & 0        & 0     & 7,063 \\
PACS Sk    & \multicolumn{3}{c}{----------~3,929~----------}  & 7  & 0        & 0     & 3,929   \\
TUB        & \multicolumn{3}{c}{---------~20,000~---------} & 250 & 0        & 0     & 17,840 \\ \hline

Total      & & & & & 947,057   & 60,821 & 80,582   \\ \hline \hline

\end{tabular}
\end{center}
\vspace{-4pt}
\caption{Number of natural images and sketches for the original datasets and for the Im4Sketch dataset partitioned into train, validation and test subsets.\label{tab:datasets2}
\vspace{-25pt}
}
\end{table}
\section{Experiments}
\label{sec:exp}

\begin{table*}[t]
\vspace{8pt}
    \centering
  \begin{minipage}[t]{.65\linewidth}
    \centering
    \footnotesize
\def\arraystretch{0.95}
\begin{center}
\setlength\extrarowheight{1pt}
\begin{tabular}{@{\xssp}l@{\xssp}l@{\xssp}l@{\xssp}l@{\xssp}l@{\xssp}l@{\xssp}l@{\xssp}l@{\xssp}l}
\hline
ID            & 0                    & 1             & 2             & 3             & 4 rBTE   & 5 rBTE   & 6            & 7    \\ \hline \hline
Geometric  & Yes                  & No            & Yes           & Yes           & Yes                   & Yes                    & Yes          & Yes   \\
Edge detector & \multirow{3}{*}{N/A} & HED           & HED           & HED           & All                   & All                    & HED          & HED      \\
NMS           &                      & Yes           & Yes           & Yes           & Yes                   & Yes                    & No           & Yes  \\
Thresholder   &                      & Fixed         & Fixed         & All           & All                   & All                    & No          & No    \\
Pretrained    & RGB/IN               & RGB/IN        & RGB/IN        & RGB/IN        & RGB/IN                & BTE/I4S                & RGB/IN       & RGB/IN  \\ \hline
Single scale & 11.4 $\pm$3.5         & 39.1 $\pm$0.8 & 47.0 $\pm$0.3 & 47.9 $\pm$0.6 & 49.7 $\pm$0.5         & 57.2 $\pm$0.5          & 5.9 $\pm$2.9 & 40.2 $\pm$0.6  \\
Multi scale  & 11.6 $\pm$3.9         & 42.4 $\pm$1.5 & 49.8 $\pm$0.6 & 50.9 $\pm$0.4 & 52.3 $\pm$0.5         & 59.6 $\pm$0.8          & 8.0 $\pm$2.2 & 41.1 $\pm$0.8  \\ \hline \hline
\end{tabular}
\end{center}

    \vspace{-4pt}
    \caption{Sketchy: ablation study for training on natural images and testing on sketches.
    \label{tab:abla}}
  \end{minipage}%
  \begin{minipage}[t]{.28\linewidth}
    \centering
    \small
\def\arraystretch{1}
\begin{center}
\setlength\extrarowheight{1pt}
\begin{tabular}{@{\xssp}l@{\xssp}c}
\hline
Method  & Photo $\longrightarrow$ Sketch \\
\hline
\textit{RGB}  & 32.6  $\pm$ 2.4 \\
\textit{SelfReg~\cite{SelfReg}}   & 33.7 $\pm$ 2.6 \\ 
\textit{L2D~\cite{Wlq+21}}   & 58.5 $\pm$ 3.9 \\ 
\textit{SagNet~\cite{SagNet}}   & 40.7 \ \ \ \ \ \ \ \  \\ 
\hline
\textit{rBTE (Ours)}   & \textbf{70.6 $\pm$ 2.2} \\
\hline \hline
\end{tabular}
\end{center}
    \vspace{-4pt}
    \caption{PACS: image-to-sketch generalization.\label{tab:tab_pacs}}
  \end{minipage}
\vspace{-20pt}
\end{table*}

In this section, experimental settings, datasets, and results are presented\footnote{The code and the proposed evaluation protocols are publicly available on \url{https://github.com/NikosEfth/im2rbte}}. Firstly, we perform an extensive set of ablations by training and testing on Sketchy dataset. Then, we train and test the proposed approach on PACS dataset to compare with prior domain generalization results. Lastly, we perform large-scale training and testing on the newly introduced Im4Sketch dataset. Recognition performance on Sketchy and Im4Sketch is evaluated by mean prediction accuracy, first estimated per class and then averaged for the whole dataset. For the comparison with the existing domain generalization methods, mean accuracy is used in order to be consistent with the literature.

\begin{table}[b]
\vspace{-12pt}
\def\arraystretch{0.95}
\begin{center}
\footnotesize
\setlength\extrarowheight{1pt}

\begin{tabular}{@{\xssp}l@{\xssp}c@{\xssp}c@{\xssp}c@{\xssp}c@{\xssp}c@{\xssp}c@{\xssp}c@{\xssp}c@{\xssp}c@{\xssp}c@{\xssp}c@{\xssp}c}
\hline
\multicolumn{1}{c}{} &
  \multicolumn{2}{c}{RGB (single)} &
  \multicolumn{2}{c}{rBTE (single)} &
  \multicolumn{2}{c}{rBTE (multi)} \\
$\arg\max$ over & all & subset & all & subset & all & subset \\ \hline
Im4Sketch     &   5.3  & 5.3   & 11.3 &  12.7 &   12.4 &  14.0 \\ \hline 
DomainNet     &    3.9 & 3.9   & 5.2  &  6.9  &   6.6  &  8.5  \\
Sketchy       &    3.9 & 12.4  & 26.0 &  42.6 &   26.8 &  43.0 \\
TUB           &   12.0 & 13.9  & 25.8 &  31.5 &   27.8 &  33.9 \\
PACS          &   11.6 & 42.2  & 24.7 &  62.7 &   23.8 &  64.8 \\\hline \hline
\end{tabular}
\end{center}
\vspace{-4pt}
\caption{Results for training on Im4Sketch. Testing is performed on the test set of Im4Sketch or its subsets that come from the different original datasets.\label{tab:global_results}
\vspace{-20pt}
}
\end{table}

\subsection{Ablation study on Sketchy}
Sketchy dataset~\cite{sbh+16} is originally created for the purpose of fine-grained sketch-based image retrieval, while we define a new evaluation protocol for our task. It consists of 125 object categories with natural images. Sketches of objects are drawn with these images as reference. The training part consists of 11,250 images and the corresponding 68,418 sketches, while the testing part consists of 1,250 images and the corresponding 7,063 sketches. Since there is no designated validation set, we randomly split the training part keeping 80\% for training and 20\% for validation.
This policy is followed both for images and sketches.

We choose the Sketchy benchmark to perform an ablation study for computational reasons.
Each ablation includes a 30 epoch training procedure followed by the evaluation; the reported numbers are averaged over five randomized runs. The results of the study are summarized in Table~\ref{tab:abla}. 
Training using single edge-maps, and fixed threshold, without geometric augmentation results in poor performance (ID=1), while simply adding geometric augmentations provides a noticeable boost (ID=2). In these two ablations the threshold is fixed and set equal to the average threshold estimated by the Otsu method on the whole dataset.
Then, using all the threshold estimation methods per image increases performance (ID=3). Using all edge-maps (ID=4) provide further boost. This variant constitutes the main approach of this work. However, instead of initializing with the result of training on ImageNet with RGB images, we also perform an experiment where the initialization is the result of training with rBTEs on Im4Sketch (described at the end of this section).
This kind of large-scale pre-training  is shown very essential (ID=5) and increases performance by 7.5\% with single scale and 7.3 in multi scale testing. Two additional ablations show that edge-map binarization is needed (ID=7 versus ID=3), and that without the NMS-based thinning the performance is very low (ID=6). The baseline approach of training on RGB images of the Sketchy dataset and then testing on sketches (ID=0) performs very poorly. This comparison demonstrates the large benefit of rBTEs for the Sketchy dataset which includes detailed and well drawn sketches.

We additionally use the Canny edge detector employed with geometric augmentations, Otsu's method to adaptively set Canny's thresholds~\cite{fyy09}, and Gaussian smoothing with $\sigma$ randomly chosen in $[1, 5]$. It achieves 44.9\% and 47.9\% accuracy in single and multi-scale testing, which is 2.1\% and 1.9\% lower than HED ablation with no adaptive threshold (ID=2).

\subsection{Single-source domain generalization comparison on PACS}

Table \ref{tab:tab_pacs} summarises the performance of our approach compared to the current state-of-the-art in domain generalization and to the baseline trained directly on RGB natural images. SelfReg~\cite{SelfReg} performs poorly as expected; it is intended  for multi-source domain generalization. 
For L2D~\cite{Wlq+21} which is designed specifically for the single-source task, we run the provided code and ensure that optimal learning rate according to validation performance on the source task is used; the reported score is reproduced.
The reported  numbers  are  averaged  over  twenty  randomized  runs. 
Our approach outperforms all other generic domain generalization methods by a large margin.

\subsection{Training on Im4Sketch and testing on all}
The proposed approach for learning without sketches is to train on the corresponding image training part of Im4Sketch by transforming them into rBTEs. In this way, both the backbone network and the soft-max classifier are trained only with rBTEs. This is performed either to obtain a sketch classifier for 874 categories, or as pre-training to obtain a better backbone network, tailored for shape-based representation (see ID=5 on Table~\ref{tab:abla}). 

There are 874 training classes in Im4Sketch, while only 393 classes have sketches for testing, since drawing all the classes is impractical at best.
Besides results at single and mutli-scale, two evaluating scenarios are reported.
First, ``argmax over-all'', is the testing over all 874 possible classes. This should be seen as an estimate of the overall performance, as the test-classes are unknown during training.
The other scenario, ``argmax over-subset'' is the testing over the classes that appears in  the sketch test set. This corresponds to an unrealistic situation, when posterior probability of classes not in the test set is known to be zero. We only report these results to provide some intuition.
The results of our Im4Sketch trained model are summarized in Table~\ref{tab:global_results} to allow future comparisons. Training on rBTEs performs significantly better than training on RGB images on all setups.

\section{Conclusions}
\label{sec:conclusions}

In this work, we are the first to train a large-scale sketch classifier that is able to recognize up to 874 categories. Due to the absence of such a large training dataset, the learning is performed without any sketches. Instead, we proposed a novel edge augmentation technique to translate natural images to a pseudo-novel domain and use it to train a network classifier. 
This tailored image-to-sketch method is noticeably better than generic single-source domain generalization approaches.

\clearpage
{\small
\bibliographystyle{IEEEtran}
\bibliography{bib}}

\end{document}